\documentclass[letterpaper, 10 pt, conference]{ieeeconf}  

\IEEEoverridecommandlockouts                              

\usepackage{amsmath} 
\usepackage{graphicx}
\usepackage{adjustbox}
\usepackage{subcaption}
\usepackage{cleveref}
\usepackage{booktabs}
\usepackage{flushend}

\title{\LARGE \bf
Finding safe 3D robot grasps through efficient haptic exploration \\with unscented Bayesian optimization and collision penalty
}

\author{Jo\~ao Castanheira$^{1}$, Pedro Vicente$^{1}$, Ruben Martinez-Cantin$^{2,3}$, Lorenzo Jamone$^{4,1}$, Alexandre Bernardino$^{1}$
\thanks{$^{1}$ Institute for Systems and Robotics, Instituto Superior T\'ecnico, Universidade de Lisboa, Lisbon, Portugal.
        {\tt\small joao.castanheira@tecnico.ulisboa.pt,  \{pvicente,alex\}@isr.tecnico.ulisboa.pt}}%
\thanks{$^{2}$ Centro Universitario de la Defensa, Zaragoza, Spain.
{\tt\small rmcantin@unizar.es}}
\thanks{$^{3}$ SigOpt Inc. San Francisco, CA}
\thanks{$^{4}$ ARQ (Advanced Robotics at Queen Mary), School of Electronic Engineering and Computer Science, Queen Mary University of London, UK
        {\tt\small  l.jamone@qmul.ac.uk}}%
\thanks{This work was partially supported by FCT (UID/EEA/50009/2013 and PD/BD/135115/2017), and by EPSRC UK (project NCNR, National Centre for Nuclear Robotics, EP/R02572X/1). 
We acknowledge the support of NVIDIA with the donation of a GPU.}
}

\usepackage{color}
\usepackage{amsmath}
\DeclareMathOperator*{\argmax}{argmax}
\usepackage{amsfonts}

\newcommand{\incubentbo}{{}\text{y}^{\text{b}\textit{opt}}_n}
\newcommand{\incubentubo}{{}\text{y}^{\text{ub}\textit{opt}}_n}

\newcommand{\optimumbo}{{}\mathbf{x}^{\text{b}\textit{opt}}_n}
\newcommand{\optimumubo}{{}\mathbf{x}^{\text{ub}\textit{opt}}_n}

\newcommand{\xoptimum}{\mathbf{x}^\textit{opt}_n}

\newcommand{\sampleMC}{y_{\text{mc}}(\xoptimum)}
\newcommand{\outcomeMC}{\Bar{\text{y}}_{\text{mc}}(\xoptimum)}
\newcommand{\stdMC}{\text{std}\Big(\text{y}_{\text{mc}}(\xoptimum)\Big)}

\newcommand{\outcomeMCubo}{\Bar{\text{y}}_{\text{mc}}(\optimumubo)}
\newcommand{\stdMCubo}{\text{std}\Big(\text{y}_{\text{mc}}(\optimumubo)\Big)}

\usepackage[utf8]{inputenc}
\usepackage{gensymb}
\usepackage{placeins}

\begin{document}
\maketitle
\thispagestyle{empty}
\pagestyle{empty}


\begin{abstract}
Robust grasping is a major, and still unsolved, problem in robotics. 
Information about the 3D shape of an object can be obtained either from prior knowledge (\textit{e.g.}, accurate models of known objects or approximate models of familiar objects) or real-time sensing (\textit{e.g.}, partial point clouds of unknown objects) and can be used to identify good potential grasps. However, due to modeling and sensing inaccuracies, local exploration is often needed to refine such grasps and successfully apply them in the real world. The recently proposed unscented Bayesian optimization technique can make such exploration 
safer by selecting grasps that are robust to uncertainty in the input space (\textit{e.g.}, inaccuracies in the grasp execution). Extending our previous work on 2D optimization, in this paper we propose a 3D haptic exploration strategy that combines unscented Bayesian optimization with a novel collision penalty heuristic to find safe grasps in a very efficient way: while by augmenting the search-space to 3D we are able to find better grasps, the collision penalty heuristic allows us to do so without increasing the number of exploration steps. 

\end{abstract}


%
\IEEEpeerreviewmaketitle

\section{Introduction}

\label{sec:intro}


Robotic grasping and manipulation has been a major research area for many years now \cite{mason85grasp, shimoga99grasp, sahbani12grasp, bohg:2014:tro}.
However, the robust grasping of arbitrary objects is still an open problem.
%
%
%
According to Bohg et al \cite{bohg:2014:tro}, an object can be (i)~known, (ii)~familiar or (iii)~unknown.
In the first category, accurate models of the object are available and can be exploited for transforming a grasping problem into a pose estimation one \cite{Huebner:2010:knowgrasp,papazov:2012:knowgrasp,ciocarlie:2014:knowgrasp}, since the optimum grasp could be learned \textit{a priori} and retrieved from a database of possible grasp poses.
In the second case, the object shares some features (\textit{e.g.}, 
visual 3D geometric features) with previous known objects and can fall back on the previous case by adjusting the final target pose with online learning methods \cite{montesano:2012:familiargrasp}.
Finally, when grasping unknown objects we should resort to real-time sensing (\textit{e.g.}, partial point clouds) using exploration for retrieving the optimum grasp pose. 

Even with known objects, it is hard to achieve a robust grasp since small errors in object sensing or motor execution may turn optimal grasps into bad grasps. Indeed, many object grasping controllers rely on a complete open-loop approach to achieve the pre-computed optimal grasp \cite{figueiredo2012,kim2014,stuckler2015,Leidner2016a,Leidner2016b}, where the robot looks once to the scene, computes the pose of the object, and then drives the arm to the grasping pose without sensory feedback. 
Due to modeling inaccuracies and sensing limitations, such open-loop approach does not permit robust grasping performance. 
As a matter of fact, very few grasping pipelines exploit tactile or visual feedback to achieve a robust grasp execution~\cite{bohg:2014:tro}. 
%
\begin{figure}[t]
    \centering
      \includegraphics[trim=0.0cm 1.05cm 0.0cm 0.5cm,clip=true,width=0.47\textwidth]{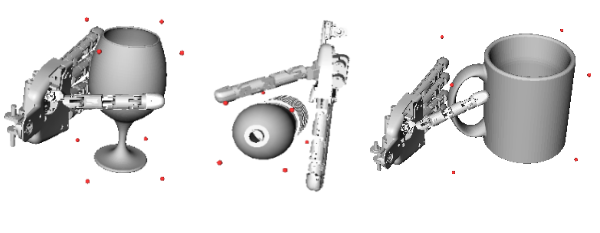}
    \caption{Examples of objects to perform grasp optimization on simulation. Initial pose for each test object.}
    \label{fig:initialpose}
\end{figure}
In general, some form of local adjustment is needed to adapt the desired grasps the objects in the real world. For instance, positioning errors can be reduced by means of visual servoing \cite{vicente:2017:icra}, or by closing the control-loop with force and torque sensors \cite{pastor:2011:iros}, or by learning the optimum grasping position with a trial-and-error approach \cite{Kroemer:2009:iros}. 



In this work, we follow the trial-and-error approach based on tactile feedback to reduce the effect of uncertainty in the input space (\textit{i.e.}, the hand positioning in 3D) and eventually obtain robust and safe grasps. We apply Bayesian optimization methods to find such grasps. This approach will minimize the number of required exploration actions needed to achieve the optimum grasp. 


%

In our previous work \cite{nogueira:iros:2016} we have introduced the unscented Bayesian optimization (UBO) algorithm to search for safe grasps in a 2D space. In this paper, we scale to 3D and prove that better grasps can be found by UBO against Bayesian optimization (BO). Notably, the superior outcome of the 3D UBO method was possible due to a new Collision Penalty (CP) strategy proposed to increases the convergence speed. 

The rest of the paper is organized as follows: we present the BO framework (Sec.~\ref{sec:bayes}), followed by the recently proposed UBO (Sec.~\ref{sec:UBO}) and we introduce the CP heuristic in Sec.~\ref{sec:CP}. Then, we present experimental results on grasping simulations in Sec.~\ref{sec:results}, using the objects presented in Fig.~\ref{fig:initialpose}, and conclude our paper in Sec.~\ref{sec:conclusions}.

\section{Bayesian optimization}
\label{sec:bayes}


In this article, we follow the BO notation presented in \cite{nogueira:iros:2016,brochu:tutbayes}. Formally, the problem is based on finding the optimum (maximum) of an unknown real valued function $f:\mathcal{X} \to \mathbb{R} $, where $\mathcal{X}$ is a compact space, $\mathcal{X} \subset \mathbb{R}^d$, and $d \ge 1$ its dimension, with a maximum budget of $N$ evaluations of the target function $f$. In our problem $f$ is a grasp quality criterion described in Sec.~\ref{sec:CP}, and $\mathcal{X}$ is the space of relative hand--object poses. For each query point $\mathbf{x_i} \in \mathcal{X}$ a grasp trial has an outcome $y_i$= $f(\mathbf{x_i})+\eta$, where $\eta$ is zero-mean noise with variance $\sigma_{\eta}^2$.

The BO consists of two stages. First, given a query point $\mathbf{x_i}$ and outcome $y_i$ we update a probabilistic surrogate model of $f$, a distribution over the family of functions $P(f)$, where the target function $f$ belongs. A very popular choice is a Gaussian process (GP), built incrementally by sampling over the input-space. Second, a Bayesian decision process, where an acquisition function uses the information gathered in the GP, is used to decide on the best point $\mathbf{x}$ to query (\textit{i.e.}, to sample) next. The goal is to guide the search to the optimum, while balancing the exploration \textit{vs} exploitation trade-off.


\subsection{Surrogate model estimations}

The Gaussian process $\mathcal{GP}(\mathbf{x}|\mu,\sigma^2,\theta)$ has inputs $\mathbf{x} \in \mathcal{X}$, scalar outputs $\text{y} \in \mathbb{R}$ and an associated kernel function $k(\cdot,\cdot)$ with hyperparameters $\boldsymbol{\theta}$. The hyperparameters are optimized during the process using the slice sampling method \cite{snoek:bayesianoptimization}, \cite{ruben:bayesopt}, resulting in $m$ samples $\mathbf{\Theta} = [\boldsymbol{\theta}_i]_{i=1}^m$.

From the GP we can get an estimate, $\hat{y}$, of  $f$ by conditioning the distribution over functions to what is already known. 
At a step $n$ we have a dataset of observations $\mathbf{\mathcal{D}}_n=(\mathbf{X},\mathbf{y})$, represented by all the queries until that step, $\mathbf{X} = (\mathbf{x}_{1:n})$, and their respective outcomes, $\mathbf{y} = (y_{1:n})$. The prediction, $y_{n+1} = \hat{y}(\mathbf{x}_{n+1})$, at an arbitrary new query point $\mathbf{x}_{n+1}$, with kernel $k_i$ conditioned on the $i$-th hyperparameter sample $k_i = k(\cdot,\cdot | \boldsymbol{\theta}_i)$, is normally distributed and given by: 
\begin{equation}
\label{eq:predictions}
\begin{split}
    &\hat{y}(\mathbf{x}_{n+1}) \sim \sum_{i=1}^m \mathcal{N}(\mu_i(\mathbf{x}_{n+1}), \sigma_i^2(\mathbf{x}_{n+1}))
    \\
    &\mu_i(\mathbf{x}_{n+1})= \mathbf{k}_i^T \mathbf{K}_i^{-1} \mathbf{y}
    \\
    &\sigma_i^2(\mathbf{x}_{n+1}) = k_i(\mathbf{x}_{n+1},\mathbf{x}_{n+1}) - \mathbf{k}_i^T \mathbf{K}_i^{-1} \mathbf{k}_i.
\end{split}
\end{equation}

The vector $\mathbf{k}_i$ is a kernel $i$ evaluated at the arbitrary query point $\mathbf{x}_{n+1}$ with respect to the dataset $\mathbf{X}$, and $\mathbf{K}_i = k_i (\mathbf{X},\mathbf{X}) + \mathbf{I}\cdot\sigma_{\eta}^2$ is the Gram matrix corresponding to kernel $k_i$ for dataset $\mathbf{X}$, with noise variance $\sigma_{\eta}^2$. Note that, because we use a sampling distribution of \mbox{\boldmath$\Theta$}, the predictive distribution at any point $\mathbf{x}$ is a mixture of Gaussians.

\subsection{Decision using acquisition function}
\label{subsec:EI}

To select the next query point at each iteration, we use the expected improvement criterion (EI) as the acquisition function. This function takes into consideration the predictive distribution for each point in $\mathcal{X}$, whose mean and variance are as in Eq. (\ref{eq:predictions}), to decide the next query point. The EI is the expectation of the improvement function $I(\mathbf{x})= \text{max}(0, \hat{y}(\mathbf{x}) - \incubentbo)$,
where the incumbent is the best outcome found until now (iteration $n$):
\begin{equation}
\label{eq:boincumbent}
    \incubentbo = \text{max}( y_{1:n} )
    .
\end{equation}
Then, the optimum value corresponds to its associated query in the dataset and is denoted as $\optimumbo$. 
Taking the expectation over the mixture of Gaussians of the predictive distribution, the EI can be computed as 
\begin{equation}
\begin{split}
    \text{EI}(\mathbf{x}) &=\mathbb{E}_{(\hat{y}|\mathbf{\mathcal{D}}_n, \boldsymbol{\theta} , \mathbf{x})} [\text{max}(\mathbf{0}, \hat{y}(\mathbf{x}) - \incubentbo]
    \\
    &= \sum_{i=1}^m [(\mu_i(\mathbf{x}) - \incubentbo )\Phi(z_i)+ \sigma_i(\mathbf{x})\phi(z_i)],
\end{split}
\end{equation}
where $z_i= (\mu_i(\mathbf{x})-\incubentbo)/\sigma_i(\mathbf{x})$, $\phi$ corresponds to the Gaussian probability density function (PDF) and $\Phi$ to the cumulative density function (CDF). Also, the pair $(\mu_i,\sigma_i^2)$ are the predictions computed in Eq. (\ref{eq:predictions}).
Then, the new query point is selected by maximizing the EI:
\begin{equation}
    \mathbf{x}_{n+1}= \argmax_{\mathbf{x} \in \mathcal{X}} EI(\mathbf{x}).
\end{equation}
Lastly, to reduce initialization bias and improve global optimality, we rely on an initial design of $p$ points based on \textit{Latin Hypercube Sampling} (LHS), as suggested in \cite{bull:LHS}.

\section{Unscented Bayesian optimization}
\label{sec:UBO}

When selecting the most interesting point to query next, acquisition functions like the EI assume that the query is deterministic. However, with input noise, our query is in fact a probability distribution.
Indeed, if we take the query's vicinity into consideration, a better notion and estimation of the expected outcome can be achieved. The size of the vicinity to be considered depends on the input noise power.

Thus, instead of analyzing the outcome of the EI to select the next query, we are going to analyze the posterior distribution that results from propagating the query distribution through the acquisition function. 
%
This can be done using the unscented transformation, a method used to propagate probability distributions through nonlinear functions with reduced computational cost and good accuracy. The unscented transform uses selected samples from the prior distribution designated sigma points, $\mathbf{x}^{(i)}$,  and calculates the value of the nonlinear function $g$ at each of these points. Then, the output distribution is computed based on the weighted combination of the transformed sigma points. 

For a $d$-dimensional input space, the unscented transformation only requires a set of $2d+1$ sigma points. If the input distribution is a Gaussian, then the transformed distribution is simply $\mathbf{x}'\sim \mathcal{N}(\sum_{i=0}^{2d} w^{(i)} g(\mathbf{x}^{(i)}), \Sigma'_{\mathbf{x}}) $, where $w^{(i)}$ is the weight corresponding to the $i^{\text{th}}$ sigma point.

The unscented transformation provides mean and covariance estimates of the new distribution that are accurate to the third order of the Taylor series expansions of $g$, provided that the original distribution is a Gaussian prior. Another advantage of the unscented transformation is its computational cost. The $2d+1$ sigma points make the computational cost almost negligible compared to other alternatives to distribution approximation. 
\subsection{Unscented expected improvement}
\label{subsec:UEI}

Considering that our prior distribution is  Gaussian $\mathbf{x} \sim \mathcal{N}(\Bar{\mathbf{x}},\mathbf{I}\sigma_x)$, then the set of $2d+1$ sigma points of the unscented transform are computed as
\begin{equation}
\label{eq:sigma}
    \mathbf{x}^0 = \Bar{\mathbf{x}}, \, 
    \mathbf{x}_{\pm}^{(i)}= \Bar{\mathbf{x}} \pm \Big( \sqrt{(d+k)\ \sigma_x}\Big)_i, \; \forall i=1...d
\end{equation}
where $(\sqrt{(\cdot)})_i$ is the i-th row or column of the corresponding matrix square root. In this case, $k$ is a free parameter that can be used to tune the scale of the sigma points. For more information on choosing the optimal values for $k$, refer to ~\cite{uhlmann:filtering}. For these sigma points, the corresponding weights are
\begin{equation}
    w^0 = \frac{k}{d+k}, \, w_{\pm}^{(i)}= \frac{1}{2(d+k)}, \,  \forall i=1...d
\end{equation}

Considering the EI as the nonlinear function $g$, then we are making a decision on the next query considering that there is input noise. This new decision can be interpreted as a new acquisition function, the unscented expected improvement (UEI). It corresponds to the expected value of the EI with respect to the input noise distribution:
\begin{equation}
    UEI(\mathbf{x}) = \sum_{i=0}^{2d} w^{(i)} EI(\mathbf{x}^{(i)}), \qquad \mathbf{x} \in \mathcal{X}.
\end{equation}

This strategy, reduces the chance that the next query point is located in an unsafe region, \textit{i.e.}, where a small change on the input (induced by noise) implies a bad outcome. 

\subsection{Unscented optimal incumbent}
  
In BO, the final decision for what we consider the optimum is independent of the acquisition function. We defined the incumbent for BO in Eq. (\ref{eq:boincumbent}), as the best observation outcome until the current iteration. However, with UEI, each query point is evaluated considering its vicinity, so as we incrementally obtain more observations and get a better GP fit, we might observe that our optimum is actually located in an unsafe region. 
Thus, instead of considering the best observation outcome as the incumbent, we also apply the unscented transformation to select the incumbent at each iteration, based on the outcome at the sigma points of each query that belongs to the dataset of observations ($\mathbf{\mathcal{D}}_n$).
Obviously, we do not want to perform any additional evaluations of $f$ because that would defeat the purpose of BO. Alternatively, we
evaluate the sigma points with our estimation $\hat{y}$, which is the GP surrogate average prediction $\mu$. Therefore, we define the unscented outcome as:
\begin{equation}
    u(\mathbf{x}) = \sum_{i=0}^{2d} w^{(i)} \sum_{j=1}^{m} \mu_j(\mathbf{x}^{(i)}), \qquad \mathbf{x} \in \mathbf{X}
\end{equation}
where $\sum_{j=1}^{m} \mu_j(\mathbf{x}^{(i)})$ is the prediction of the GP according to Eq. (\ref{eq:predictions}), integrated over the kernel hyperparameters and at the sigma points of Eq. (\ref{eq:sigma}). Under these conditions, the incumbent for the UBO is defined as $\incubentubo = u(\optimumubo)$,
where $\optimumubo = \argmax_{\mathbf{x}_i \in \mathbf{x}_{1:n}} u(\mathbf{x}_i)$ is the optimal query until that iteration according to the unscented outcome. For further information on the performance of the UBO on synthetic functions, refer to~\cite{nogueira:iros:2016}. 



\section{Grasp quality metric and collision penalty}
\label{sec:CP}
In our problem the target function $f$ is the quality of a given relative hand--object pose. Different grasp quality metrics have been defined in the literature \cite{roa:graspmetric}, usually based on the contact points, torques and forces applied to the object. 
In our work, we use the Grasp Wrench Volume metric, introduced in \cite{miller:1999:graspmetric} and defined as
$f = \text{Volume} (\mathcal{P})$,
where $\mathcal{P}$ is the convex hull of the \textit{Grasp Wrench Space}~\cite{pollard:1996:gws,borst:1999:gws}.
%
In our context, starting the hand with an initial pose $\mathbf{x}$ (as in Fig.~\ref{fig:initialpose}), the fingers are closed until touching the object, and the grasp quality metric is then computed.

We assume that approximate information about the object size and location is available, and is used to limit the exploration space. However, there are configurations that result in unfeasible grasps where the robot's hand collides with the object, even before attempting to close the hand. 
This indicates that the problem has constraints. Although there has been some recent work on BO with constraints \cite{gardner14:2014:constrained,gelbart:2014:constrained}, we opted for the simpler approach of adding a penalty as described in \cite{MartinezCantin18cyb}. This approach means that the input space remains unconstrained, improving the performance of the acquisition function's optimization. Additionally, due to kernel smoothing, we also get a safety area around the collision query where the function is only partly penalized.

Other research works on robotic grasping optimization had different approaches to deal with collisions. Some skip the collision query (\textit{e.g.}, \cite{allen:graspplanning}) and others give it a grasp quality of zero (\textit{e.g.},  \cite{nogueira:iros:2016}). In the first case, by ignoring the query we are losing information about the target function, hence not reducing uncertainty. In the second case, although uncertainty is reduced, the zero value is usually associated to a pose where the hand does not touch the object at all. Therefore, by modeling a collision with a zero value, one cannot distinguish a collision from a non-contact.
Instead, we propose a penalization factor that will drive the search away from collision locations, ensuing a reduction of explored area and consequently leading to faster convergence. 


We propose a Collision Penalty (CP) that accounts for the level of penetration of the hand in the object. The CP is calculated by finding the number of joints in the robot's hand that collide with the object, $n_j \in \mathbb{N}$, which indicates a measure of penetration in the object:
\begin{equation}
\label{eq:CP}
    \text{CP}(n_j)=1-  \mathrm{e}^{-\lambda n_j},
\end{equation}
where $\lambda$ is a tuning parameter used to smooth the penalty.
The CP is an heuristic used only to improve the convergence speed. Therefore, during the optimization we redefine the target function as $f' = f + \text{CP}$. Note that, in the evaluation process, we still resort to the original $f$.

\begin{figure} []
    \centering
    \begin{subfigure}{0.47\textwidth}
        \includegraphics[width=0.98\textwidth]{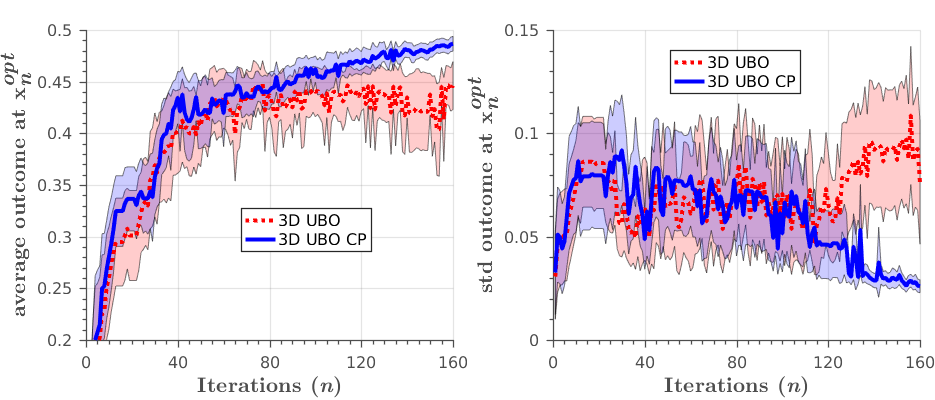}
        \caption{\textbf{Glass: CP vs $\overline{\text{CP}}$} (UBO 3D).} 
        \label{fig:glassCPvsnoCP}
    \end{subfigure}
        \begin{subfigure}{0.47\textwidth}
        \includegraphics[width=0.98\textwidth]{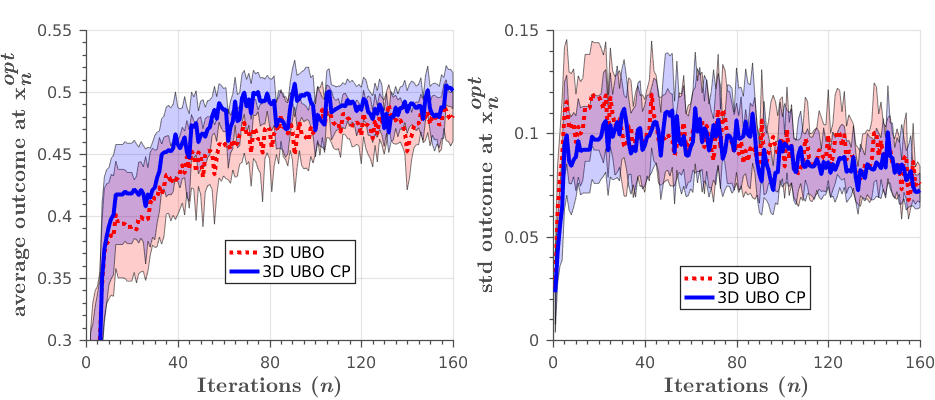}
        \caption{\textbf{Bottle: CP vs $\overline{\text{CP}}$} (UBO 3D).} 
        \label{fig:bottleCPvsnoCP}
    \end{subfigure}
        \begin{subfigure}{0.47\textwidth}
        \includegraphics[width=0.98\textwidth]{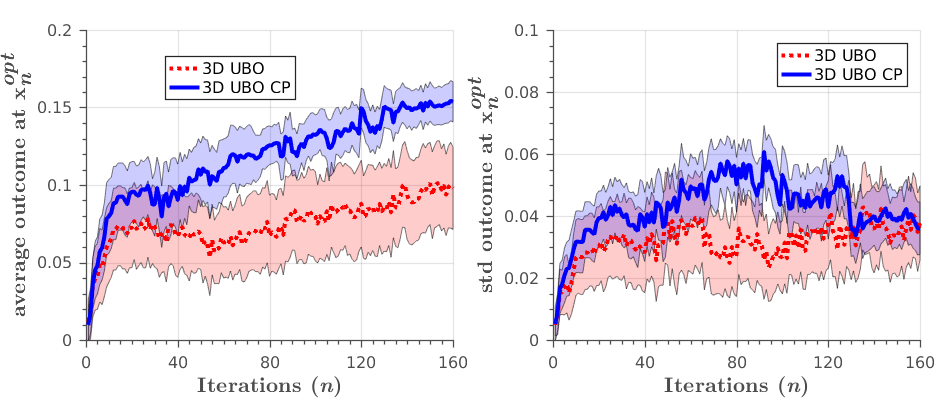}
        \caption{\textbf{Mug: CP vs $\overline{\text{CP}}$} (UBO 3D).} 
        \label{fig:mugCPvsnoCP}
    \end{subfigure}
    \caption{\textbf{CP vs $\overline{\text{CP}}$} (UBO 3D). Left: expected outcome of current optimum $\outcomeMCubo$, Right: Variability of the outcome $\stdMCubo$. Best seen in color.}
\end{figure}

\section{Implementation}
\label{sec:imp}
All the results were obtained from simulations using the Simox toolbox~\cite{vahrenkamp:2013:simox}.
This toolbox simulates the iCub's hand grasping task arbitrary objects. At the beginning of each experiment, we define an initial pose for the hand and a motion trajectory for the finger joints. The hand is placed parallel to one of the object's facets with the thumb aligned with one of the neighbor facets. This defines the canonical pose of the hand with respect to the object. Then, at each trial (optimization step) a new hand pose is defined with respect to the canonical pose by incremental translations: ($\delta_x, \delta_y, \delta_z$). All other parameters (\textit{e.g.}, hand orientation, finger joints trajectories) are identical for all trials. In 2D optimization, we optimize ($\delta_x, \delta_y$), while $\delta_z$ is kept fixed. In 3D, we optimize ($\delta_x, \delta_y, \delta_z$). For dimensions $x$ and $y$, the bounds are set to the object's dimensions, as for $z$, which is the approach direction, the bounds extend from the surface of the object's facet to the plane where the hand is no longer able to touch the object.

At each trial, after the hand is placed in a new pose, the fingers joints move according to predefined motion trajectories until they touch the object. The trajectories are set so that the hand performs a power grasp on the object, in which all fingers are closed at the same time, following a movement synergy defined in previous work \cite{bernardino13synergies}. 
When the fingers motion is finished, the grasp quality metric  defined in Sec.~\ref{sec:CP} is computed. 
If a collision between the hand (either palm or fingers) and the object is detected when positioning the hand in the new pose, the CP is applied, and the grasping motion will not be executed. For the computation of the CP, the tuning parameter is set to $\lambda=0.1$.

The \textit{BayesOpt} library \cite{ruben:bayesopt} is used to perform both BO and UBO. 
In the GP we use the Matérn kernel with $\upsilon = 5/2$. The input noise at each query point is assumed to be white Gaussian, $\mathcal{N}(0,I\sigma_{\mathbf{x}})$, with $\sigma_{\mathbf{x}} = 0.03$ (note that the input space was normalized in advance to the unit hypercube $[0,1]^d$). We assume the grasp quality metric to be stochastic, due to small simulation errors and inconsistencies, thus we set $\sigma_{\eta} = 10^{-4}$. 

\begin{figure}
    \centering
    \begin{subfigure}{0.47\textwidth}
        \includegraphics[width=0.98\textwidth]{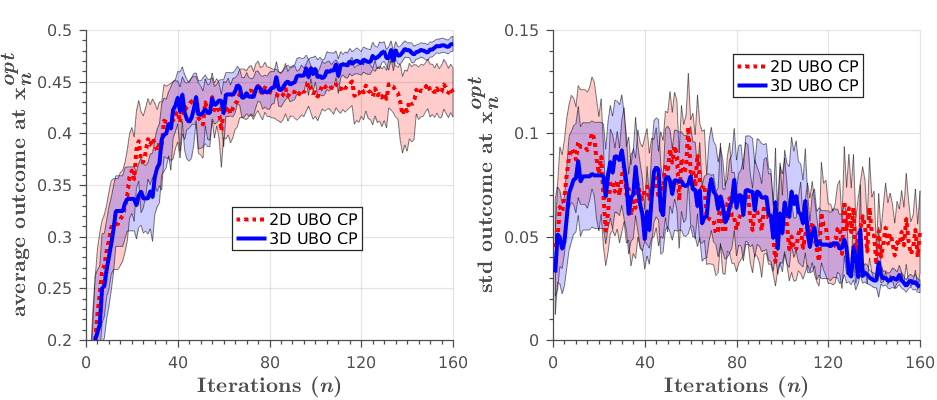}
        \caption{\textbf{Glass: 2D vs 3D} (UBO CP).}
        \label{fig:glass2dvs3dCP}
    \end{subfigure}
    \begin{subfigure}{0.47\textwidth}
        \includegraphics[width=0.98\textwidth]{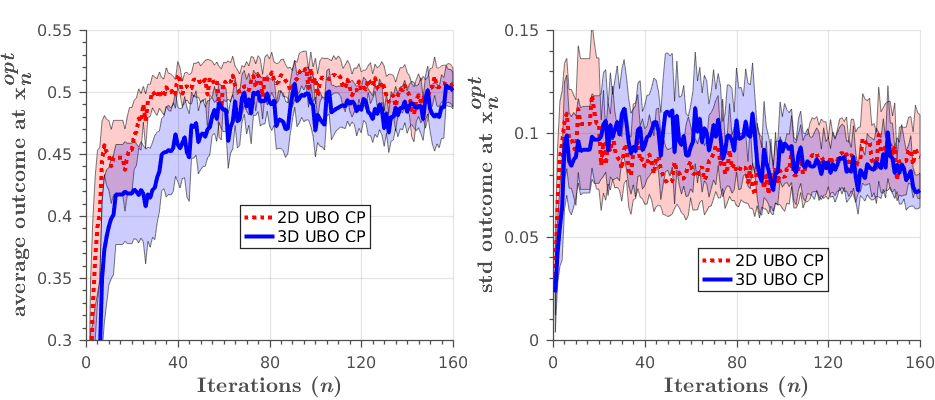}
        \caption{\textbf{Bottle: 2D vs 3D} (UBO CP)}
        \label{fig:bottle2dvs3dCP}
    \end{subfigure}
    \begin{subfigure}{0.47\textwidth}
        \includegraphics[width=0.98\textwidth]{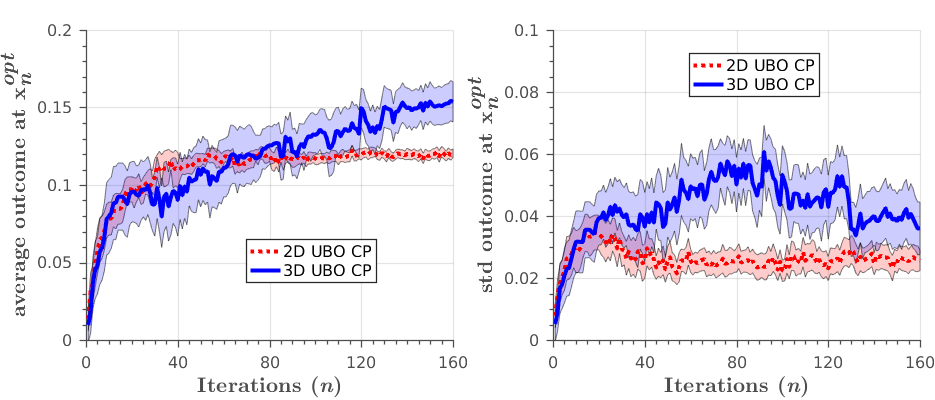}
        \caption{\textbf{Mug: 2D vs 3D} (UBO CP).}
        \label{fig:mug2Dvs3DCP}
    \end{subfigure}
\caption{\textbf{2D vs 3D} (UBO CP). Left: expected outcome of current optimum $\outcomeMCubo$, Right: Variability of the outcome $\stdMCubo$. Best seen in color.}
\end{figure}

\section{Experimental results}
\label{sec:results}

In this section, we start by describing the experiments performed to evaluate the benefits of CP (Sec.~\ref{subsec:CPvsnoCP}); then we compare UBO in 2D and 3D search problems (Sec.~\ref{subsec:2Dvs3D}); and lastly we compare UBO with BO in the 3D case (Sec.~\ref{sucsec:UBOvsBO}). 



To reproduce the effect of the input noise, we obtain Monte Carlo samples at the optimum in each iteration, $\sampleMC$, according to the input noise distribution $\mathcal{N}(0,I\sigma_{\mathbf{x}})$. Remember that $\xoptimum$ corresponds to $\optimumbo$ when performing BO, and to $\optimumubo$ for UBO. By analyzing the outcome of the samples we can estimate the expected outcome from the current optimum $\outcomeMC$, and the variability of the outcome $\stdMC$. These metrics allow us to assess if the optimum belongs to a safe region. Indeed, if $\outcomeMC$ decreases over time (which cannot occur in classical BO) it should be correlated to the fact that the optimum ($\optimumbo$) is inside an unsafe area and is not a robust grasp.



For each experiment, we performed 20 runs of the robotic grasp simulation for all test objects. The robot hand posture for each object is initialized as shown in Fig.~\ref{fig:initialpose}. Every time a new optimum is found, we collect 10 Monte Carlo samples at its location to get $\sampleMC$. Each run starts with 20 initial iterations using LHS, followed by 140 iterations of optimization. The shaded region in each plot represents a 95\% confidence interval. All the quantitative results from each experiment, at its last iteration, are presented in Table~\ref{tab:results}.


\subsection{Benefits of CP}
\label{subsec:CPvsnoCP}

To assess the benefits of CP, we performed two types of experiments for each object, a 3D UBO with and without CP. As we can see in Fig.~\ref{fig:glassCPvsnoCP} and \ref{fig:bottleCPvsnoCP}, the addition of CP to the optimization process provides a boost in convergence speed for both the glass and the bottle. Also, by penalizing collisions we are reducing the regions that are worth exploring, meaning that the robot is actually able to find a better grasp at the end, both in terms of mean and variance.

The mug is the most challenging object to learn, since the optimization is performed on the mug's facet that includes the handle. In 3D, the handle is inside the search-space, leading to a large number of configurations that result in collisions, consequently undermining the convergence to the optimum. This is a situation where the CP really thrives. By penalizing these collisions, we are driving our search away from the inside of the handle and finding a safer grasp outside the handle. 
The results in Fig.~\ref{fig:mugCPvsnoCP} show how dramatic the improvement is, achieving higher mean values with great confidence level (\textit{i.e.}, a smaller shaded region).

\begin{figure}[]
    \centering
    \begin{subfigure}{0.47\textwidth}
        \includegraphics[width=0.98\textwidth]{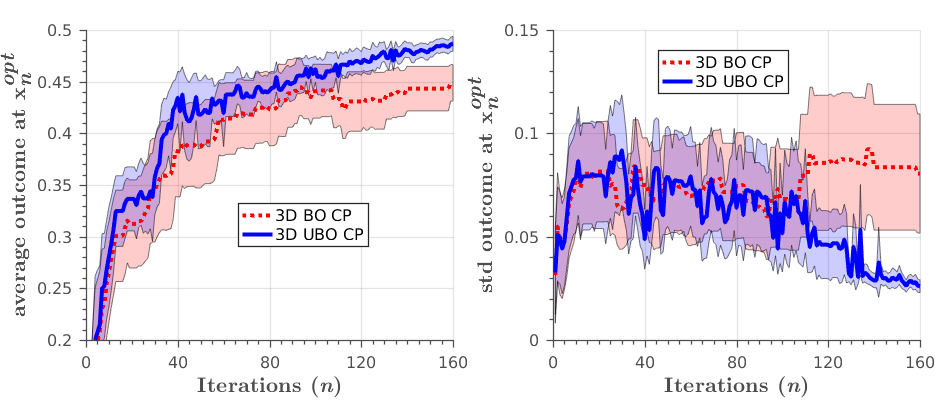}
        \caption{\textbf{Glass: BO vs UBO} (3D CP).}
        \label{fig:glassBovsUBOCP}
    \end{subfigure}
    \begin{subfigure}{0.47\textwidth}
        \includegraphics[width=0.98\textwidth]{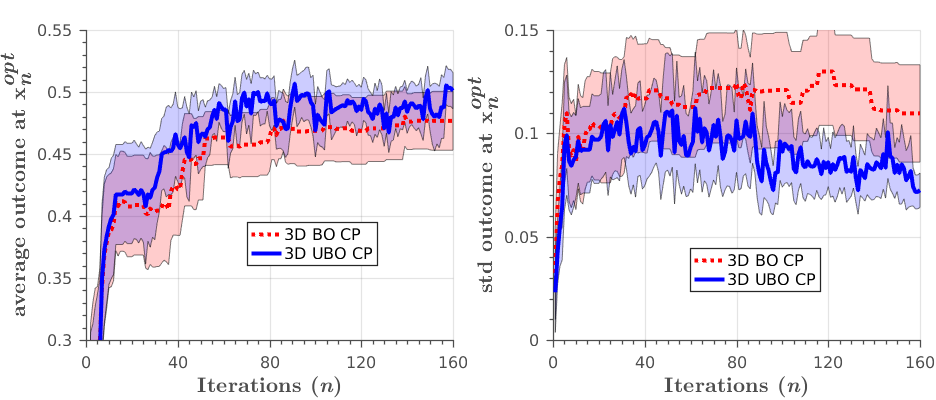}
        \caption{\textbf{Bottle: BO vs UBO} (3D CP).}
        \label{fig:bottleBOvsUBOCP}
    \end{subfigure}
    \begin{subfigure}{0.47\textwidth}
        \includegraphics[width=0.98\textwidth]{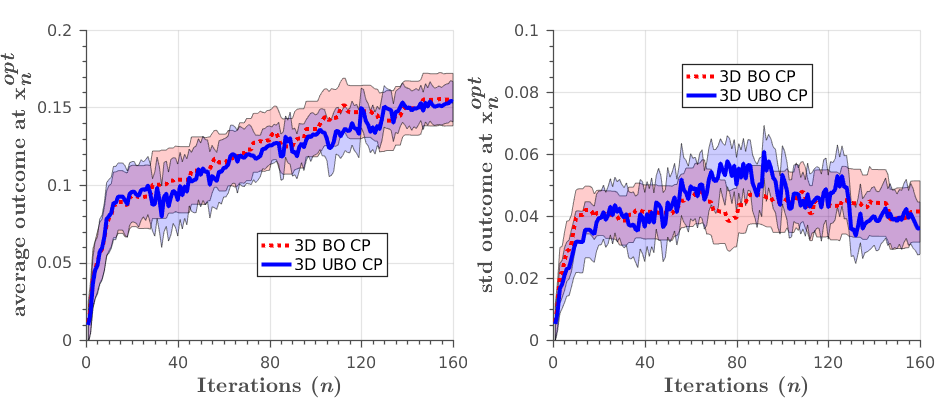}
        \caption{\textbf{Mug: BO vs UBO} (3D CP).}
        \label{fig:mugBOvsUBOCP}
    \end{subfigure}
    \caption{\textbf{BO vs UBO} (3D CP). Left: expected outcome of current optimum $\outcomeMC$, Right: Variability of the outcome $\stdMC$. Best seen in color.}
\end{figure}

\subsection{Generalization to 3D}
\label{subsec:2Dvs3D}

We performed UBO with CP in both 2D and 3D to provide evidence that UBO generalizes well into a higher dimension space, \textit{i.e.}, in 3D we only need a few extra evaluations to reach the same results obtained in 2D. 

In Fig.~\ref{fig:glass2dvs3dCP}, we observe that for the glass, even though 2D reaches better mean values right after the learning starts (iteration 20), 3D is able to reach the same level around iteration 40 and proceeds to surpass it achieving better results. As for the bottle, in Fig.~\ref{fig:bottle2dvs3dCP}, the mean value of the 3D case trails almost the whole process, only edging out the 2D results close to the end of the budget.

In the mug, Fig.~\ref{fig:mug2Dvs3DCP}, the 3D optimization only manages to reach similar mean values at around iteration 65. As explained in Sec.~\ref{subsec:CPvsnoCP}, this is due to the high amount of queries that result in collisions when we are optimizing in a 3D search-space. 


We must point out that the $z$ coordinate in the 2D optimization was chosen to ensure a fair comparison with the 3D, setting it to the parallel plane where the optimal grasp should be.
However, the better results obtained for both glass and mug in 3D indicate that the optimum $z$ was somewhere else. As we can observe in Fig.~\ref{fig:bestubo2dcp} and Fig.~\ref{fig:bestubo3dcp}, the visual difference between the best grasps in 2D and 3D is not noticeable, even though 3D still achieves better results.
Therefore, the generalization to the 3D search-space is arguably needed since a better grasp point was found during the 3D optimization. On the bottle object, the initial $z$ coordinate was set closer to the optimum. 

\begin{table}[]
    \Large
    \centering
    \vspace*{0.0cm}
    \begin{adjustbox}{max width=0.48\textwidth}
    \begin{tabular}{l *{6}{c}}
      \toprule

      {} & \multicolumn{2}{c}{Glass} & \multicolumn{2}{c}{Bottle} & \multicolumn{2}{c}{Mug}  \\
      \cline{2-7}
            & $\outcomeMC$ & $\stdMC$ & $\outcomeMC$ & $\stdMC$ & $\outcomeMC$ & $\stdMC$ \\
      \midrule
      2D UBO CP & 0.4396 & 0.0536 & \textbf{0.5026} & 0.0887 & 0.1205 & \textbf{0.0256}\\
      3D UBO $\overline{\text{CP}}$ & 0.4462 & 0.0761 & 0.4810 & 0.0754 & 0.0979 & 0.0378 \\
      3D UBO CP  & \textbf{0.4867} & \textbf{0.0260} & 0.5011 & \textbf{0.0725} & \textbf{0.1567} & 0.0361\\
      3D BO CP  & 0.4489 & 0.0805 & 0.4767 & 0.1097 & 0.1551 & 0.0415\\
      \bottomrule
    \end{tabular}
    \end{adjustbox}
    \caption{Results at the last iteration ($n=160$) of the optimization process (means over all runs).}
    \label{tab:results}
\end{table}


\subsection{Advantages of UBO over BO}
\label{sucsec:UBOvsBO}

Here we compare UBO against BO in 3D using CP, to conclude if the advantages of UBO described in \cite{nogueira:iros:2016} in 2D, generalize to 3D. 

\begin{figure} []
    \centering
    \begin{subfigure}{0.15\textwidth}
        \includegraphics[trim=0.0cm 0.6cm 0.0cm 0.8cm,clip=true,width=0.9\textwidth]{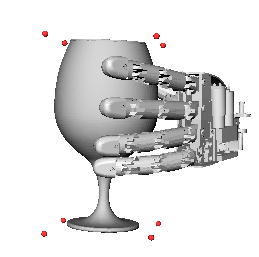}
        \caption{UBO 2D CP}
        \label{fig:bestubo2dcp}
    \end{subfigure}
    \begin{subfigure}{0.15\textwidth}
        \includegraphics[trim=0.0cm 0.6cm 0.0cm 0.8cm,clip=true,width=0.9\textwidth]{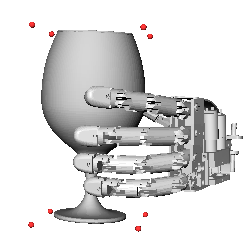}
        \caption{BO 3D CP}
        \label{fig:bestbo3dcp}
    \end{subfigure} 
    \begin{subfigure}{0.15\textwidth}
        \includegraphics[trim=0.0cm 0.70cm 0.0cm 0.8cm,clip=true,width=0.9\textwidth]{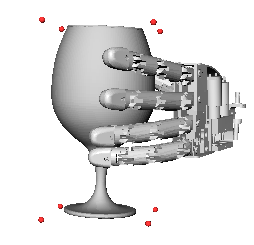}
        \caption{UBO 3D CP}
        \label{fig:bestubo3dcp}
    \end{subfigure}
    
    \caption{Best grasps in one of the runs.
    The best grasp of UBO \textbf{2D} CP (a) is similar to the UBO \textbf{3D} CP (c). The best grasp achieved by\textbf{ BO} is in an unsafe zone (b). The \textbf{UBO}'s best grasp is more robust to input noise (c). Check Table~\ref{tab:bestgrasps} for the grasp metrics in these configurations. }
    \label{fig:bestgrasps}
\end{figure}

The results collected from both methods show that we are still able to learn safer grasps using UBO in 3D. The advantage is clear for both the glass, Fig.~\ref{fig:glassBovsUBOCP}, and bottle, Fig.~\ref{fig:bottleBOvsUBOCP}, where the UBO achieves higher mean values and lower variance. In the mug, Fig.~\ref{fig:mugBOvsUBOCP}, we get competitive mean values using BO, but UBO finds an optimum with lower variance.
The visual comparison between the two optimization strategies (BO and UBO) is displayed in Fig.~\ref{fig:bestbo3dcp} and ~\ref{fig:bestubo3dcp}, where we can observe the best grasps achieved in one of the 20 runs.


\begin{table}[]
        \centering
            \begin{tabular}{l c c c}
            \toprule
                          & \textbf{UBO 2D CP}  & \textbf{BO 3D CP}   & \textbf{UBO 3D CP}  \\
            \midrule
            $\incubentbo$ and $\incubentubo$   &  $ 0.467$   & $0.649$    &  $0.497$   \\
            $\outcomeMC$  &  $ 0.4424$  & $ 0.383$   &  $0.472$   \\
            $\stdMC$      &  $ 0.0471$  & $0.207$   &  $0.0203$  \\
            \bottomrule
            \end{tabular}
        \caption{The grasp metric corresponding to the Monte Carlo sampled hand configurations shown in Fig.~\ref{fig:bestgrasps}.}
        \label{tab:bestgrasps}
\end{table}

\section{Conclusions}
\label{sec:conclusions}

This work has validated the application of Unscented Bayesian Optimization to 3D grasp optimization. We show that it outperforms the classical Bayesian Optimization in this problem and generalizes well from the existing results in 2D search to a more challenging 3D search, without compromising the optimization budget. We propose a collision penalty function to force the search algorithm away from potential collision configurations, thus speeding up the convergence of the method. In future work we will study how to extend the method to the full 6D (translation+rotation) optimization and the application of the method in a real robotic anthropomorphic hand with 3D force sensors in the finger’s phalanges \cite{paulino:2017:icra}.





\bibliographystyle{IEEEtran}
\bibliography{bib}
%



\end{document}